\documentclass[letterpaper]{article} 
\usepackage{aaai2026}
\usepackage{times}  
\usepackage{helvet}  
\usepackage{courier}  
\usepackage[hyphens]{url}  
\usepackage{graphicx} 
\usepackage{natbib}  
\usepackage{caption} 
\usepackage{makecell}
\usepackage{booktabs}
\usepackage{algorithm}
\usepackage{multirow}

\usepackage{newfloat}
\usepackage{listings}
\DeclareCaptionStyle{ruled}{labelfont=normalfont,labelsep=colon,strut=off} 
\floatstyle{ruled}
\newfloat{listing}{tb}{lst}{}
\floatname{listing}{Listing}
\usepackage{amssymb}
\usepackage{amsmath}
\usepackage{algorithm}
\usepackage{algpseudocode}
\nocopyright 

\title{DTRNet: Dynamic Token Routing Network to Reduce Quadratic Costs in Transformers}

\definecolor{mypurple}{RGB}{128,0,128}

\author{
    Aman Sharma$^{1}$, \quad
    Saeed Najafi$^{2}\thanks{Work done while at Huawei.}$, \quad
    Parsa Farinneya $^{1}$, \quad
    Benyamin Jamialahmadi $^{1}$, \\
    Marzieh S. Tahaei $^{1}$, \quad
    Yuhe Fan $^{1}$, \quad
    Mehdi Rezagholizadeh $^{4}\footnotemark[1]$, \quad
    Boxing Chen $^{1}$, \quad 
    Aref Jafari $^{1, 3}$ \quad
}
\affiliations{
    $^{1}$Huawei Noah’s Ark Lab \hspace{0.3cm}
    $^{2}$University of Alberta \hspace{0.3cm}
    $^{3}$University of Waterloo \hspace{0.3cm}
    $^{4}$Advanced Micro Devices, Inc. (AMD) \\
    aman.sharma4@h-partners.com \hspace{0.3cm} aref.jafari@uwaterloo.ca,

}

\usepackage{bibentry}

\begin{document}
\pagestyle{plain}
\maketitle

\begin{abstract}

Transformers achieve state-of-the-art results across many tasks, but their uniform application of quadratic self-attention to every token at every layer makes them computationally expensive. 
We introduce DTRNet (Dynamic Token Routing Network), an improved Transformer architecture that allows tokens to dynamically skip the quadratic cost of cross-token mixing while still receiving lightweight linear updates. By preserving the MLP module and reducing the attention cost for most tokens to linear, DTRNet ensures that every token is explicitly updated while significantly lowering overall computation. This design offers an efficient and effective alternative to standard dense attention.
Once trained, DTRNet blocks routes only ~10\% of tokens through attention at each layer while maintaining performance comparable to a full Transformer. It consistently outperforms routing-based layer skipping methods such as MoD and D-LLM in both accuracy and memory at matched FLOPs, while routing fewer tokens to full attention. Its efficiency gains, scales with sequence length, offering significant reduction in FLOPs for long-context inputs. By decoupling token updates from attention mixing, DTRNet substantially reduces the quadratic share of computation, providing a simple, efficient, and scalable alternative to Transformers. 
\footnote{The code is released at \textcolor{blue}{https://github.com/Aman26Sharma/DTRNet}}

\end{abstract}


\section{Introduction}
Transformers have emerged as the dominant architecture across natural language processing (NLP), vision, and multi-modal tasks, powering state-of-the-art (SOTA) performance in large language models (LLMs) such as GPT-4, Claude, and Gemini \cite{openai2023gpt4,claude,reid2024gemini}. Despite their empirical success, the quadratic complexity of self-attention poses a significant bottleneck, particularly as sequence lengths and model sizes scale beyond billions of parameters~\cite{transformer}. This limitation has motivated a growing body of work focused on improving the efficiency of Transformers through adaptive computation, token routing, and sparse attention mechanisms~\cite{elhoushi2024layerskip,wu2024layer,raposo2024mixture,deeprenetrouting,sparseattention}.

A key observation in Transformer-based models is that not all tokens require equal compute at every layer \cite{shin2025orthoranktokenselectionsink}. This is supported further by our layerwise cosine similarity analysis of token embeddings in a dense Transformer (See Section Methodology), which reveals that inner-layer embeddings change only marginally across adjacent layers, highlighting significant redundancy in token updates across layers. These findings suggest that uniformly applying attention to every token at every layer may be inefficient, and that routing-based mechanisms can yield substantial savings by skipping unnecessary computations.


Recent work has explored reducing Transformer computation through adaptive token routing. Mixture-of-Depth (MoD) \cite{raposo2024mixture} applies expert-choice routing on alternating layers, selectively choosing which tokens are processed while skipping the rest. In contrast, D-LLM \cite{NEURIPS2024_dllm} performs token-choice gating at every layer, dynamically deciding whether each token should pass through a Transformer block or bypass it entirely.
Despite their differences, both approaches share a key limitation: tokens that bypass a block skip both the self-attention and MLP components, receiving no update. This results in lower representational quality and weaker performance compared to standard Transformers. 



Inspired by recent work, highlighting the relative importance of MLP/FFN sublayers and the redundancy of many attention components \cite{dong2025attention,cao2024head,yu2023neuron}, we introduce the \textbf{Dynamic Token Routing Network (DTRNet)}, a Transformer architecture that retains MLP and reduces the quadratic computation via a two-path design. At each layer, a learned router assigns each token to either (i) a standard \emph{quadratic} full-attention path or (ii) a lightweight \emph{linear} path that updates the token using shared value and output projections. 
Unlike prior routing-based methods (e.g., MoD, D-LLM) that skip both attention and MLP entirely, our DTR layers retains the MLP for all tokens.




Once trained, DTRNet routes only $\sim$10\% of tokens on average through attention at each layer, while achieving performance comparable to a full Transformer. 
DTRNet consistently outperforms both MoD and D-LLM in terms of accuracy and memory under same computation budget. DTRNet achieves this while routing fewer tokens to full attention compared to MoD and D-LLM, thereby further reducing the quadratic complexity of the Transformer. The advantage of DTRNet becomes even more pronounced as sequence lengths increase, providing significantly lower FLOPS  and lower perplexity than the baselines. These findings highlight the effectiveness of DTRNet’s projection-based update path, which not only preserves representational quality but also enables a substantial reduction in quadratic computation. This makes DTRNet particularly well-suited for language modeling and long-context applications, where sequences can grow far beyond the training horizon.

Our main contributions are:
\begin{itemize}
\item 
A dynamic routing architecture that enable explicit token updates at reduced cost via  a lightweight linear path and a quadratic attention path while retaining a shared MLP block.
\item A simple training objective that penalizes routing tokens to attention, allowing the model to learn when attention is needed while preserving accuracy.
\item Empirical analysis showing that DTRNet outperforms SOTA performance at comparable compute and improves efficiency across model scales (360M, 1.3B). Our evaluations show that efficiency and FLOPs gain of DTRNet increases with sequence length, surpassing dense Transformers and prior routing methods.
\end{itemize}




\section{Related Work}

\paragraph{Sparse Attention Architectures.}
The quadratic complexity of self-attention in Transformers has spurred research into more efficient architectures. Sparse attention mechanisms, such as those in Longformer \cite{beltagy2020longformer}, Reformer \cite{kitaev2020reformer} and BASED~\cite{arora2024simple}, reduce computation by limiting attention to local windows or approximating it with kernel-based methods. Unlike these approaches, which modify attention patterns, DTRNet focuses on token-level computation, selectively routing tokens to bypass quadratic attention.

\paragraph{Depth-Adaptive Inference.}
Depth-adaptive methods aim to reduce inference costs by dynamically adjusting the number of layers processed per input. Methods like LayerSkip \cite{elhoushi2024layerskip}, SortedNet \cite{valipour2023sortednet}, SortedLLaMA~\cite{kavehzadeh2024sorted} and Balcony \cite{jamialahmadi2025balcony} employ different variants of early-exit strategies, enabling intermediate layers to produce predictive outputs. Flextron \cite{cai2024flextron} uses a router to select sub-networks of varying depths. These methods, however, apply uniform computation across all tokens in a sequence, limiting granularity. Skipping more than 30\% of layers often degrades performance due to insufficient token-specific adaptivity \cite{jamialahmadi2025balcony}. DTRNet overcomes this by routing tokens individually, allowing fine-grained computation allocation tailored to each token’s needs.

\paragraph{Token-Level Routing.}
Token-level depth-wise routing offers a more granular approach by dynamically allocating computation per token. 
Mixture-of-Depths (MoD) \cite{zhou2022mixture} introduces a learned router at each layer that selects a top-$k$ subset of tokens for full attention and MLP computation, while the rest are forwarded through residual connections without updates. To support inference-time deployment, an auxiliary network is trained to predict routing decisions. However, this introduces a consistency mismatch between training and inference, as the router behavior changes post-training. This often results in suboptimal token selection and degraded performance, especially in large-scale models.

D-LLM \cite{dllm2024} proposes dynamic token-wise layer skipping for pretrained Transformers by attaching lightweight token routers at each layer. These use Gumbel-Softmax sampling to make discrete routing decisions, with auxiliary losses encouraging partial usage of each layer. While effective at reducing computation, D-LLM suffers from layer starvation, where certain layers, particularly deeper ones, receive limited gradient signal, leading to underutilized capacity and weaker generalization. This issue is especially pronounced in long-context scenarios.

\begin{figure}[th]
    \centering
    \includegraphics[width=1.11\linewidth]{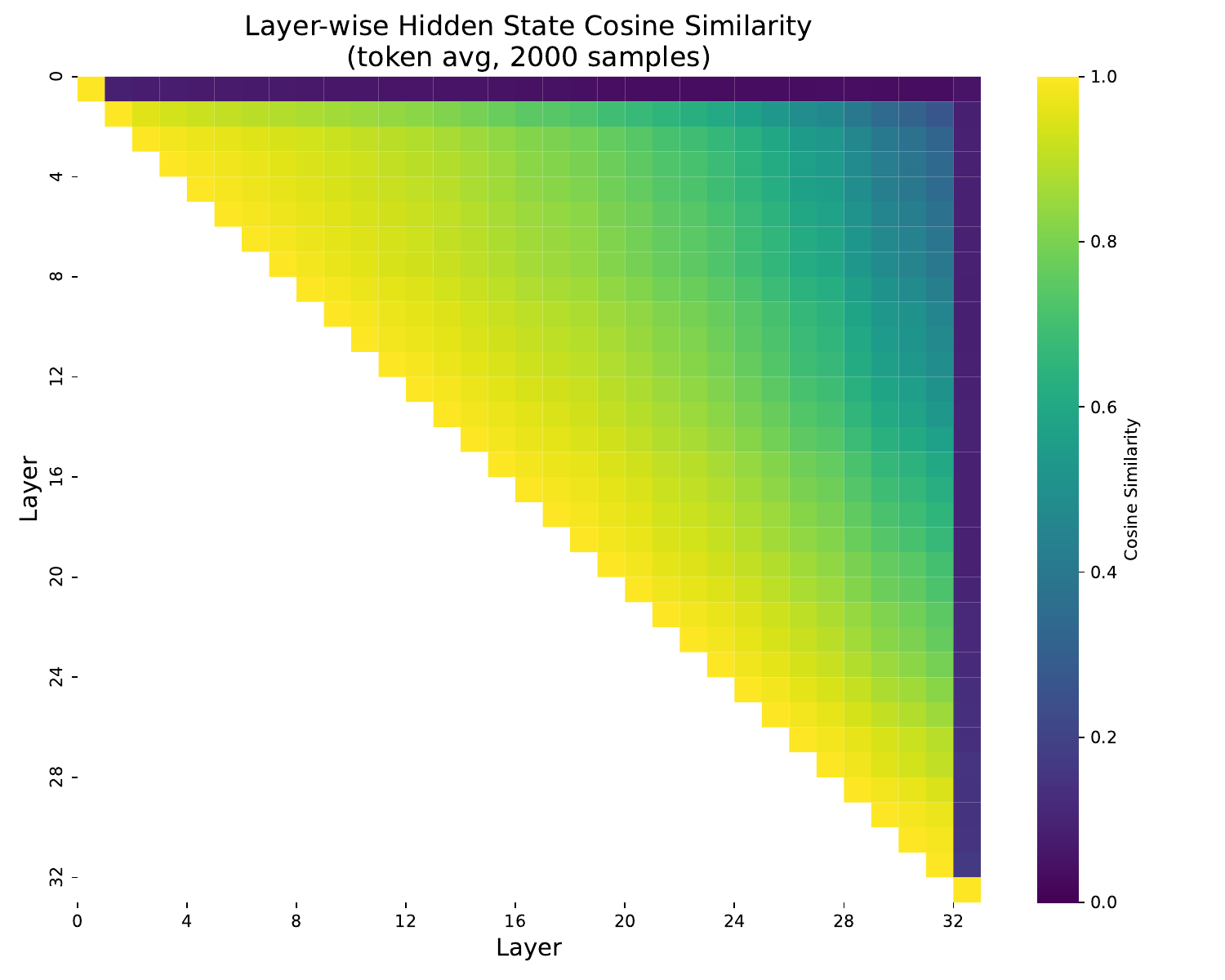}
    \caption{Average Layerwise cosine similarity of token embeddings in a 1.3B SmolLM model}
    \label{fig:layer_layer_cosine_sim}
\end{figure}

\section{Methodology}
\label{sec:methodology}
DTRNet is a Transformer architecture designed to reduce the quadratic complexity of self-attention by dynamically routing tokens through either a full attention path or a lightweight projection-based path. Motivated by empirical evidence of computational redundancy in Transformer token updates, DTRNet ensures all tokens receive meaningful updates while minimizing attention overhead. This section outlines the motivation from token redundancy analysis, the DTRNet layer structure, training strategy, and architectural design choices.

\subsection{Motivation: Token Redundancy}
\label{motivation}
To inform DTRNet’s design, we analyzed token representation redundancy in a 1.3B-parameter Transformer (SmolLM \url{https://huggingface.co/blog/smollm}) evaluated on WikiText \cite{merity2016pointer}. We computed the average cosine similarity between token embeddings \( x^{(l)} \in \mathbb{R}^{n \times d} \) across layers \( l \), where \( n \) is the sequence length and \( d \) is the hidden dimension.
The resulting similarity matrix \( S \in \mathbb{R}^{L \times L} \), visualized in Figure~\ref{fig:layer_layer_cosine_sim}, reveals high similarity between adjacent inner layers (e.g., \( S_{i,i+1} \approx 0.98 \)), indicating minor representation changes, while the first and last layers show lower similarity with their neighbors. These findings suggest  redundancy in adjacent layers showing potential for reduced computation. 
In contrast, the first and last layers exhibit low similarity with their neighbors, indicating more substantial transformations at the boundaries.

\begin{figure}[t]
    \centering
    \includegraphics[width=1.05\columnwidth, trim=1cm 1cm 0cm 2cm, clip]{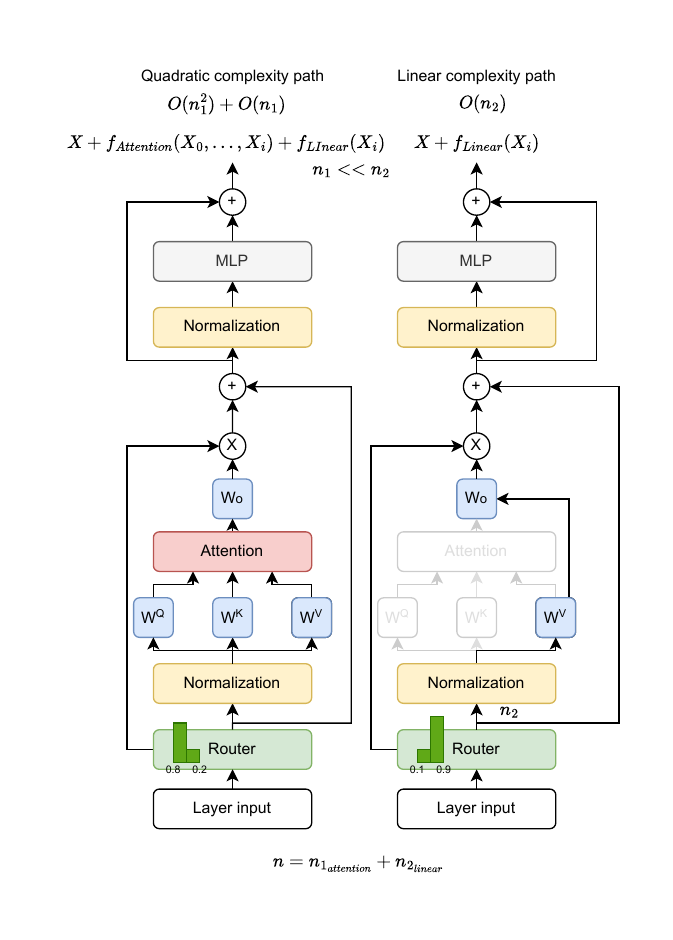}
    \caption{\textbf{DTRNet Layer.} Left: tokens routed to the self-attention path undergo full cross-token mixing. Right: tokens routed to the projection-only (bypass) path skip mixing and receive a token-local update via $W^V$ and $W^O$, followed by the shared FFN. Both paths share parameters}
    \label{fig:DTRNET}
\end{figure}

\subsection{DTRNet Layer Structure}

Each DTRNet layer operates on the input token embeddings 
\( X^{(l)} = [x^{(l)}_1, \dots, x^{(l)}_n] \in \mathbb{R}^{n \times d} \), 
where \( n \) is the number of tokens and \( d \) is the hidden dimension, and $l$ is the layer index. 
A learned \emph{router} \( \mathcal{G}^{(l)} \) processes each token embedding \( x^{(l)}_i \) 
and outputs a probability distribution over two computational paths:
\begin{itemize}
    \item the \textbf{quadratic path}, which applies standard multi-head self-attention, and
    \item the \textbf{linear path}, which applies linear projections without cross-token interaction.
\end{itemize}

\paragraph{Token Router.}
The router \( \mathcal{G}^{(l)} \) is a two-layer feedforward network:
\begin{equation}
\mathcal{G}^{(l)}_i = \text{softmax}\left(\text{SiLU}(x^{(l)}_i W_1^{(l)} ) \cdot W_2^{(l)}\right) \in \mathbb{R}^2
\end{equation}
where \( W_1^{(l)} \in \mathbb{R}^{d \times \frac{d}{2}} \), \( W_2^{(l)} \in \mathbb{R}^{\frac{d}{2} \times 2} \), 
and the output \( \mathcal{G}^{(l)}_i = [g^{(l)}_{i,\text{attn}}, g^{(l)}_{i,\text{bypass}}] \) 
represents the soft routing scores.\\

We perform \textit{hard routing} by selecting the path with the higher score:
\begin{equation}
\delta^{(l)}_i = 
\begin{cases}
1 & \text{if } g^{(l)}_{i,\text{attn}} > g^{(l)}_{i,\text{bypass}} \\
0 & \text{otherwise}
\end{cases}
\end{equation}

\paragraph{Computation Paths.}
Each token is routed to one of two computational branches:
\\

\textbf{Quadratic Path:} If \( \delta^{(l)}_i = 1 \), the token is updated via standard multi-head self-attention:
\begin{equation}
y^{(l)}_i = \text{MLP}\left(g^{(l)}_{i,\text{attn}} \cdot \text{Attn}(x^{(l)}_i)\right)
\end{equation}
where:
\begin{equation}
    \begin{split}
         & \text{Attn}(x^{(l)}_i) = \left( \text{softmax}\left( \frac{q_i K^\top}{\sqrt{d}} + M_{\text{C}} \right) V \right) W^O \\
        & q_i= x^{(l)}_i W^Q, \quad K = X^{(l)} W^K, \quad V = X^{(l)} W^V
    \end{split}
\end{equation}
and  \( M_{\text{C}} \in \mathbb{R}^{n \times n} \) is the causal mask.\\

\textbf{Linear Path:} If \( \delta^{(l)}_i = 0 \), the token undergoes a self-only linear transformation:
\begin{equation}
y^{(l)}_i = \text{MLP}\left(g^{(l)}_{i,\text{bypass}} \cdot x^{(l)}_i W^V W^O\right)
\end{equation}
The projection weights \( W^Q, W^K, W^V, W^O \in \mathbb{R}^{d \times d} \) and $MLP(.)$ are shared across both computation paths.
This design ensures that \emph{all tokens are updated}, with the router dynamically choosing whether full attention is necessary. \\

Although routing decisions are hard at inference time, we use the soft scores 
\( g^{(l)}_{i,\text{attn}} \) and \( g^{(l)}_{i,\text{bypass}} \) to weight the outputs of 
both paths during training. This allows gradients to flow into the router 
parameters via backpropagation. This strategy is commonly used in many dynamic routing mechanisms. \cite{shazeer2017outrageously, fedus2022switch, raposo2024mixture}. 

\paragraph{Sparse Attention Equivalence}
The hard-routing scheme induces a dynamic sparse attention mechanism. Let \( \tilde{g}^{(l)} = [\delta^{(l)}_1, \dots, \delta^{(l)}_n] \in \{0,1\}^n \) denote the binary vector of routing decisions for layer \( l \). The effective attention mask is:
\begin{equation}
M^{(l)} = \tilde{g}^{(l)} \cdot \tilde{g}^{(l)\top} \in \{0,1\}^{n \times n}
\end{equation}
This mask restricts attention to interactions among tokens routed to the attention path, creating an input-dependent sparsity pattern. Unlike static sparse attention methods (e.g., Longformer \cite{beltagy2020longformer}, DTRNet’s sparsity adapts to token importance, reducing quadratic complexity to linear for \~90\% of tokens on average (as validated in experiments). This equivalence highlights DTRNet’s ability to achieve efficiency comparable to sparse attention models while maintaining flexibility through learned routing.



\subsection{Training and Regularization}

To encourage sparse attention usage, we train DTRNet with
a composite loss. Let $g^{(l)}_{i,\text{attn}} \in [0,1]$ denote the router’s soft score for sending token $i$
to attention at layer $l$, and let $\delta^{(l)}_i \in \{0,1\}$ be the corresponding hard
decision ($1$ = routed to attention). Define the per‑layer \emph{attention load}, the
number of tokens not skipped as:
\[
f_l^{\text{att}} \;=\; \sum_{i=1}^n \delta^{(l)}_i,
\]
where $n$ is the sequence length. Inspired by MoE load balancing, but specialized to a
single “expert” (the attention path), we penalize the product of the load and the
aggregate attention scores:
\begin{equation}
\mathcal{L} \;=\; \mathcal{L}_{\text{CE}} \;+\; \lambda \sum_{l=1}^L {\alpha_l} \,\big\| G^{(l)}[:,0] \big\|_1,
\end{equation}
where $\alpha_l = \frac{f_l^{\text{att}}}{\sum_{i=1}^L f_i^{\text{att}}}$ defines a layer-wise normalized attention load, to encourage balanced and efficient attention usage across layers. $\mathcal{L}_{\text{CE}}$ is the task cross‑entropy loss, $L$ is the number of
layers, $\lambda > 0$ controls the strength of the routing penalty, $G^{(l)} \in
\mathbb{R}^{n \times 2}$ stacks the router’s per‑token scores for the two paths at
layer $l$, and $G^{(l)}[:,0] = [g^{(l)}_{1,\text{attn}}, \dots, g^{(l)}_{n,\text{attn}}]$.
The $\ell_1$ norm $\|\cdot\|_1$ aggregates attention mass across tokens. This objective
discourages over‑routing by increasing the penalty when many tokens are sent to
attention or when their attention probabilities are large, thereby encouraging sparse,
budget‑aware use of attention while preserving task performance.

\subsection{Architectural Design Choices}
The redundancy analysis motivates two key design choices. First, the high similarity in inner layers justifies the bypass path, which uses shared projections (\( W^V, W^O \)) to update tokens at linear cost, avoiding the full skipping of layers as in MoD or D-LLM. Second, the distinct behavior of boundary layers necessitates full-attention layers at the first and last positions to handle input adaptation and output alignment. To preserve long-range dependencies, we interleave DTRNet layers with full Transformer layers (e.g., T-D-T or T-D-D-T patterns), ensuring periodic cross-token mixing. Ablation studies (see Appendix A2) confirm that these configurations  optimize the trade-off between efficiency and accuracy.

\begin{table*}[!ht]
\centering
\resizebox{\textwidth}{!}{%
\begin{tabular}{lccccccccccccc}
\toprule
\textbf{Model} & \makecell{\textbf{FLOPs} \\ \textbf{Ratio}} & \makecell{\textbf{WIKI} \\ \textbf{ppl (↓)}} & \makecell{\textbf{LMBD} \\ \textbf{ppl (↓)}} & \makecell{\textbf{LMBD} \\ \textbf{Acc (↑)}} & \makecell{\textbf{ARC-C} \\ \textbf{Acc (↑)}} & \makecell{\textbf{ARC-E} \\ \textbf{Acc (↑)}} & \makecell{\textbf{BQ} \\ \textbf{Acc (↑)}} & \makecell{\textbf{HLSWG} \\ \textbf{Acc (↑)}} & \makecell{\textbf{PIQA} \\ \textbf{Acc (↑)}} & \makecell{\textbf{T-MMLU} \\ \textbf{Acc (↑)}} & \makecell{\textbf{WG} \\ \textbf{Acc (↑)}} & \makecell{\textbf{AVG} \\ \textbf{Acc (↑)}} \\
\midrule
\multicolumn{13}{c}{\textbf{360M}} \\
\midrule
SmolLM                           & 1.00 & 30.18 & 52.59 & 30.93 & 24.32 & 55.72 & 59.88 & 31.80 & 65.72 & 32.62 & 52.88 & 44.23 \\
D-LLM                     & 0.84 & 32.82 & 63.31 & 30.12 & 22.18 & 54.80 & 60.55 & 31.32 & 63.33 & 31.97 & 48.62 & 42.86 \\
MoD                      & 0.84 & 32.16 & 50.00 & 31.48 & 22.27 & 54.46 & 61.74 & 31.17 & 64.80 & 27.72 & 51.46 & 43.13 \\
DTRNet Trilayer (Ours)                 & 0.78 & 31.75 & 64.00 & 29.19 & 23.04 & 55.56 & 61.62 & 31.21 & 66.05 & 31.79 & 51.14 & 43.70 \\
DTRNet Bilayer (Ours)            & 0.84 & 30.68 & 56.52 & 31.22 & 23.46 & 56.06 & 59.39 & 31.73 & 66.38 & 32.51 & 54.14 & \textbf{44.36} \\
\midrule
\multicolumn{13}{c}{\textbf{1.3B}} \\
\midrule
SmolLM                       & 1.00 & 17.44 & 15.61 & 44.67 & 33.28 & 70.24 & 61.96 & 42.02 & 72.96 & 28.93 & 58.17 & 51.53 \\
D-LLM                     & 0.85 & 18.40 & 17.83 & 43.01 & 31.91 & 68.98 & 60.24 & 40.99 & 71.87 & 29.96 & 58.01 & 50.62 \\
MoD                      & 0.84 & 19.38 & 18.11 & 42.64 & 33.11 & 68.06 & 56.09 & 41.18 & 71.76 & 29.31 & 55.56 & 49.71 \\
DTRNet Bilayer (Ours)            & 0.85 & 17.95 & 16.69 & 43.80 & 32.68 & 69.15 & 60.15 & 41.11 & 71.16 & 32.72 & 57.38 & \textbf{51.02} \\
\bottomrule
\end{tabular}%
}
\caption{Comparison of DTRNet and baselines at both 360M and 1.3B scales. WIKI and LMBD are reported in perplexity (lower is better), while all other metrics are accuracy scores (higher is better). FLOPs ratio is calculated relative to the dense baseline.}
\label{tab:dtrnet_main_table}
\end{table*}

\begin{figure*}[ht]
    \centering
    \includegraphics[width=1\linewidth]{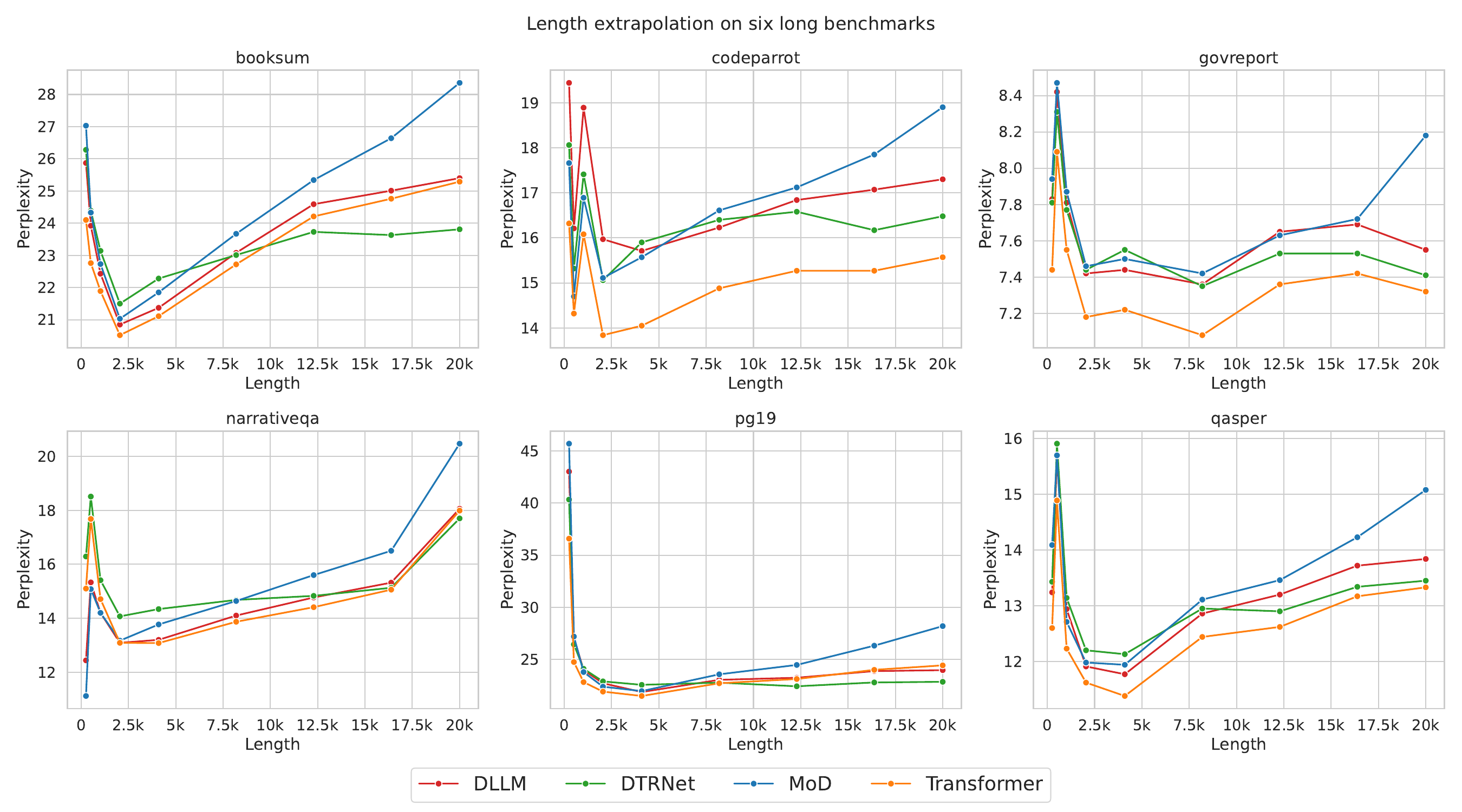}
    \caption{Perplexity across increasing sequence lengths for LongLM benchmarks on 6 different tasks. DTRNet maintains lower perplexity than MoD and D-LLM and matches or outperforms the dense Transformer on tasks like BookSum, NarrativeQA, and PG-19, demonstrating strong extrapolation to long-context inputs.}
    \label{fig:ppl_analysis}
\end{figure*}
\begin{figure}[t]
    \centering
    \includegraphics[width=1.1\linewidth,trim={1.8cm 0 0 0}, clip]{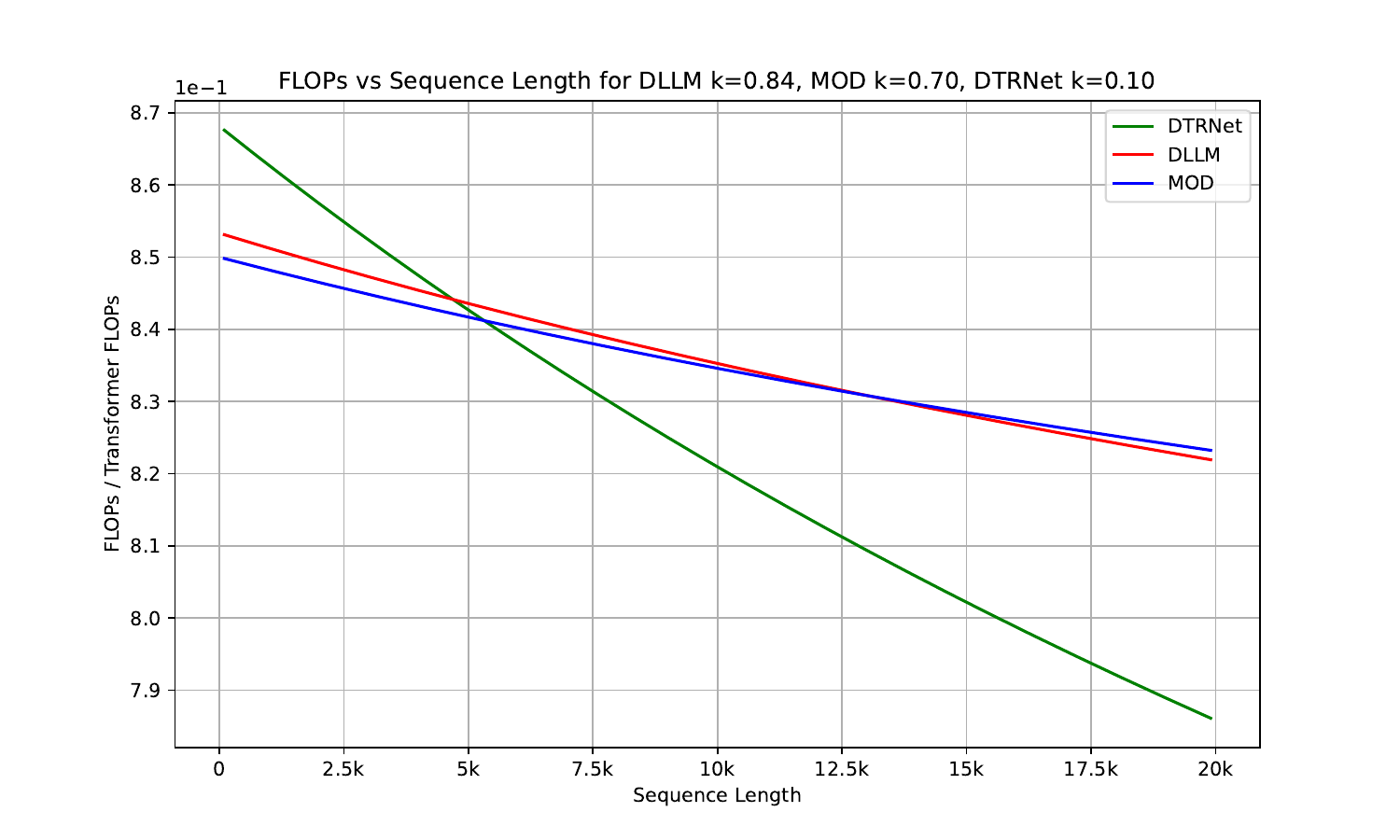}
    \caption{Theoretical FLOPs comparison as sequence length increases. DTRNet shows improved efficiency over MoD, D-LLM, and baseline Transformers for long inputs.}
    \label{fig:flops_vs_seqlen}
\end{figure}
\begin{figure*}
    \centering
    \includegraphics[width=1.05\linewidth]{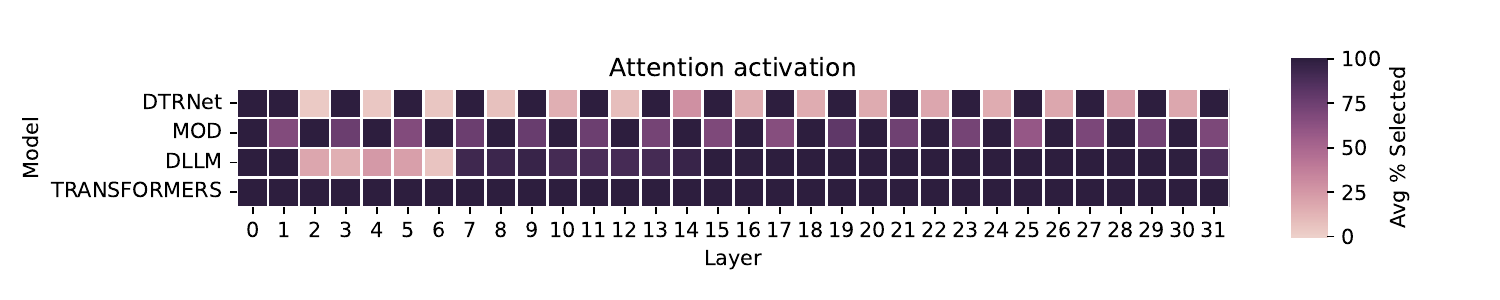}
    \caption{Average Percentage of tokens going to attention per layer for D-LLM, MoD and DTRNet. We observe much less tokens are going into attention in DTRNet while maintaining performance}
    \label{fig:layer_activation}
\end{figure*}
\begin{figure}
    \centering
    \includegraphics[width=1\linewidth]{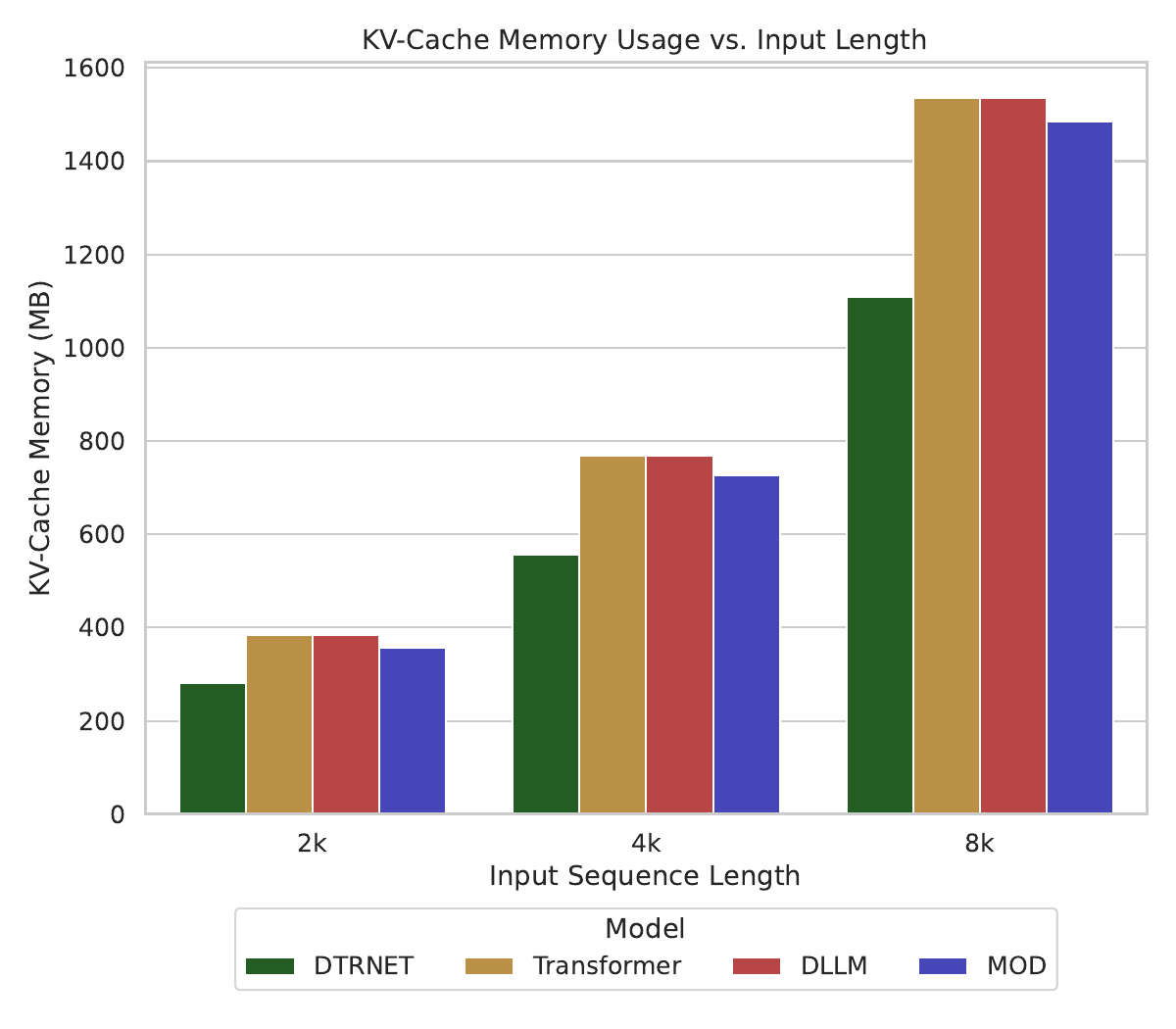}
    \caption{KV cache memory for different sequence lengths}
    \label{fig:kv_memory}
\end{figure}

\section{Experiments}

Our experiments encompass a comprehensive comparison against the following recent state-of-the-art baselines: SmolLM, MoD, D-LLM. These experiments are conducted using both 360M-parameter and 1.3B-parameter model variants. To ensure fairness, all models are trained under identical conditions.

\subsection{Training Setup}

We consider two model scales, 360M and 1.3B parameters.

\textbf{360M Models:} These follow the SmolLM-360M architecture, consisting of 32 Transformer layers, a hidden dimension of 960, and an intermediate feedforward dimension of 2560. Training is done on 15B tokens randomly sampled from the FineWeb-Edu corpus \cite{penedo2024fineweb}, using a global batch size of 384. \\

\textbf{1.3B Models:} These models consist of 24 layers, with a hidden size of 2048 and an intermediate size of 5632. They are trained on 100B tokens sampled from the FineWeb-Edu dataset \cite{penedo2024fineweb}, with a global batch size of 1024.\\

All models are trained with a maximum sequence length of 2048 tokens. Optimization is performed using the AdamW optimizer \cite{loshchilov2017decoupled} with a peak learning rate of 3e-4, weight decay of 0.01, and gradient clipping set to 0.1. The learning rate schedule follows cosine decay with a warmup ratio of 0.1. For all variants, we employ the LLaMA 2 tokenizer \cite{touvron2023llama2}, with a vocabulary size of 32,000.

All models are trained under a fixed FLOPs budget at 2k sequence length to ensure that observed performance differences can be attributed to architectural and routing design rather than computational disparity. Architecture and  architecture-specific hyperparameters of DTRNet and other baselines are as follows: \\

\textbf{DTRNet.}  
We explore multiple architectural configurations for integrating DTRNet blocks into Transformer models. In all variants, the first and last layers are standard Transformer layers. The main configurations are as follows:
\begin{itemize}
    \item \textbf{DTRNet-BiLayer}: Inserts one DTRNet block after each Transformer block. (T-D-T-D-T-D)
    \item \textbf{DTRNet-TriLayer}: Inserts two DTRNet blocks after each Transformer block. (T-D-D-T-D-D)
\end{itemize}

For training, we set the regularization strength $\lambda$ to $8 \times 10^{-4}$ for the 360M parameter model and $6 \times 10^{-4}$ for the 1.3B parameter model. To enable efficient batch training, we employ the variable-length attention function \texttt{flash\_attn\_varlen\_func()} from FlashAttention 2 \cite{dao2023flashattention2}.\\

\textbf{MoD.}  
We adopt the bi-layer routing configuration from the original MoD paper, placing one MoD block after each Transformer layer. We reproduce the classifier for inference with a linear layer. The top-$k$ token routing ratio is fixed at $0.7$ to maintain comparable compute (FLOPs) with other baselines for a fair evaluation.\\

\textbf{D-LLM.}  
For D-LLM, we set the acceleration rate $\Omega = 0.85$ and the auxiliary loss coefficient $\alpha = 1.0$, following the recommended values in the original work to have same computational budget. The router used is two layer MLP router based on original paper. The architecture retains full Transformer layers in the first two layers, while all subsequent layers are replaced with D-LLM blocks. The first two tokens of each sequence are reserved and consistently passed through the Transformer in every layer as done in original D-LLM setup. Unlike the original D-LLM setup, which applies LoRA \cite{hu2021lora} to the query/key/value projections and fine-tunes the routing components, our setup performs full model pretraining from scratch and hence does not use any parameter-efficient fine-tuning techniques.




\subsection{Evaluation Benchmarks}

To comprehensively evaluate DTRNet, we conduct experiments across both standard language understanding tasks and long-context extrapolation benchmarks.

\paragraph{Standard Language Understanding Benchmarks.}
To assess general language understanding, reasoning, and factual recall capabilities, we evaluate our models on a broad set of standardized benchmarks. These benchmarks span diverse tasks such as multiple-choice question answering, commonsense reasoning, and cloze-style completion. Specifically, we include evaluations on ARC (Easy and Challenge) \cite{clark2018arc}, BoolQ (BQ) \cite{clark2019boolq}, HellaSwag (HLSWG) \cite{zellers2019hellaswag}, PIQA \cite{bisk2020piqa}, Tiny MMLU (T-MMLU) \cite{hendrycks2020mmlu}, Winogrande (WG) \cite{sakaguchi2021winogrande}, LAMBADA (LMBD) \cite{paperno2016lambada} and WIKITEXT (WIKI) \cite{merity2016pointer}. All the evaluations on these tasks are done in zero-shot setting.

\paragraph{Length Extrapolation Benchmarks.}
In addition to standard evaluation, we assess the model's ability to extrapolate to long sequences. Following prior work, we test on sequences of up to 20K tokens using six long-context benchmarks. These tasks test long-range dependency modeling, memory retention, and contextual reasoning under extreme sequence lengths. Experiments are conducted for 1.3B models for these benchmarks. We apply YaRN \cite{peng2023yarn} with a factor of 10.0 to extrapolate the sequence length. Performance on these benchmarks is reported in terms of perplexity.

\paragraph{Memory Efficiency Analysis.}
In addition to task performance and extrapolation capabilities, we analyze the memory efficiency of various architectures across different sequence lengths. This analysis is crucial for understanding the practical deployment implications of each model, especially under resource-constrained settings. These analysis are performed for 360M models.

\section{Results and Analysis}

\label{sec::results}

\paragraph{Standard Language Understanding Benchmarks.}
Table~\ref{tab:dtrnet_main_table} presents the performance of our models on a diverse suite of standard language understanding tasks. At the 360M scale, our best-performing variant \textbf{DTRNet-BiLayer} achieves the highest overall average accuracy (44.36), surpassing both the MoD baseline (43.13) and the D-LLM baseline (42.86). This demonstrates the effectiveness of our sparse routing design under a matched computational budget across models. Notably, DTRNet-BiLayer performs on par with the dense Transformer baseline (SmolLM: 44.23), even having better performance in some of the tasks such as LAMBADA, ARC-E, PIQA and WG, indicating that accuracy can be preserved while significantly reducing compute. DTRNet Trilayer configuration also outperforms MoD and D-LLM on average while having less FLOPs. These findings support the analysis that adjacent Transformer layers exhibit high token similarity, enabling effective token skipping across layers without substantial performance degradation.

At the 1.3B scale, DTRNet again demonstrates competitive performance, achieving an average accuracy of 51.02, exceeding both MoD (49.71) and D-LLM (50.62), and closely approaching the dense Transformer baseline SmolLM (51.53). This trend is further supported by perplexity results on WIKI and LAMBADA, where DTRNet achieves lower perplexity than both sparse baselines and performs comparably to the full Transformer with reduced FLOPs.


\paragraph{Length Extrapolation Benchmarks.}
As shown in Figure~\ref{fig:ppl_analysis}, DTRNet consistently achieves lower perplexity than both MoD and D-LLM across all evaluated long-context tasks, particularly at extreme sequence lengths (e.g., 20K tokens). Notably, DTRNet matches or outperforms the dense Transformer baseline in several tasks, specifically \textit{BookSum} \cite{kryscinski2021booksum}, \textit{NarrativeQA} \cite{kovcisky2018narrativeqa}, and \textit{PG-19} \cite{rae2019pg19}, demonstrating strong generalization to longer sequences.

Importantly, these improvements come with greater computational efficiency. Figure~\ref{fig:flops_vs_seqlen} shows the relative FLOPs ratio to dense Transformer as sequence length increases. DTRNet exhibits a more rapid decline in FLOPs compared to MoD and D-LLM, as it routes significantly fewer tokens through the quadratic attention path. At a 20K sequence length, DTRNet operates at a FLOPs ratio of 0.785, whereas both MoD and D-LLM maintain higher ratios around 0.82. Despite this reduced computational cost, DTRNet still outperforms both sparse baselines, and is at par with dense Transformer, highlighting its effective token selection strategy and strong extrapolation capability.

\paragraph{Memory Efficiency Analysis.}
Figure~\ref{fig:layer_activation} illustrates the percentage of tokens routed to the attention mechanism during inference at each layer for the DTRNet and baseline models. DTRNet passes only 10\% tokens to attention per layer, and this selection remains remarkably uniform across all DTRNet layers. This consistent sparsity is a direct result of the weighted regularization introduced during training, which encourages balanced token flow across layers and prevents overreliance on specific depths.
In contrast, D-LLM lacks an explicit load balancing mechanism, leading to an imbalanced token distribution. As seen in the figure, the initial D-LLM layers route very few tokens to attention, an effect also observed during training, which results in starvation of these layers and subsequently degrades overall performance.
MoD, on the other hand, maintains a more uniform token flow, with approximately 70\% of tokens being routed through attention across layers. However, this behavior is driven by expert choice routing during training and replicated at inference time using a learned classifier. This indirect mechanism limits MoD’s ability to adapt to dynamic token importance, potentially impacting its effectiveness.

Passing fewer tokens to attention not only reduces computational cost but also significantly lowers memory usage, particularly in the context of key-value (KV) caching during autoregressive inference. As shown in Figure~\ref{fig:kv_memory}, DTRNet consistently consumes substantially less memory for KV cache compared to other baselines under matched FLOPs budgets. This advantage becomes increasingly noticeable as the sequence length grows, highlighting DTRNet’s suitability for long-context inference.
While D-LLM proposes a KV eviction strategy to mitigate memory usage, its implementation involves masking the KV cache during attention computation. Consequently, this method does not reduce the actual KV cache footprint, and as observed in Figure 6, its memory usage remains comparable to that of a dense Transformer. In contrast, DTRNet achieves true memory savings by avoiding KV allocation for unselected tokens entirely, making it a more practical and scalable solution for memory-efficient inference.



These results underscore the effectiveness of DTRNet in balancing compute and performance, offering strong length extrapolation capabilities while maintaining or improving performance with reduced quadratic attention overhead.






\section{Conclusion}
We introduced DTRNet, a dynamic token routing architecture that reduces Transformer inference cost by decoupling attention from token updates. By replacing full-layer skipping with a linear path that retains the MLP, DTRNet ensures that all tokens receive meaningful updates, even when attention is bypassed. This lightweight linear path enables substantial compute savings while maintaining accuracy. Through extensive experiments, we showed that DTRNet consistently outperforms SOTA approaches MoD and D-LLM in both accuracy and memory at matched FLOPs, particularly on long-context tasks. Our work highlights the utility of projection-based updates and fine-grained token routing in building efficient, high-performance language models, paving the way for scalable deployment of LLMs in latency-sensitive and low memory settings.
Despite these promising results, DTRNet introduces additional training complexity due to the dynamic routing mechanism, which may require careful hyperparameter tuning and stabilization techniques. 
Future work includes training DTRNet with larger models and longer input sequences, as well as extending the approach to multimodal architectures and encoder-decoder settings to explore its applicability in more complex scenarios.


\bibliography{aaai2026}




\clearpage 
\appendix
\section{Appendix}

\subsection{A1. Expert Choice vs Token Choice Routing}

\begin{table}[h]
\centering
\resizebox{\linewidth}{!}{%
\begin{tabular}{lccc}
\toprule
\textbf{Benchmark} & \makecell{\textbf{Smollm} \\ \textbf{360M}} & \makecell{\textbf{DTRNet} \\ \textbf{Expert Choice}} & \makecell{\textbf{DTRNet} \\ \textbf{Token Choice}} \\
\midrule
FLOPs       & 1.00 & 0.87 & 0.85  \\
ARC-C Acc (↑)      & 24.32  & 23.81  & 23.46  \\
ARC-E Acc (↑)     & 55.72  & 55.98  & 56.06  \\
BQ Acc (↑)        & 59.88  & 57.00  & 59.39  \\
HLSWG Acc (↑)      & 31.80  & 31.66  & 31.73  \\
LMBD Acc (↑)       & 30.93  & 30.71  & 31.22  \\
PIQA Acc (↑)      & 65.72  & 65.61  & 66.38  \\
T-MMLU Acc (↑)    & 32.62  & 26.80  & 32.51  \\
WG Acc (↑)      & 52.88  & 51.30  & 54.14  \\
\midrule
AVG Acc (↑)        & 44.23  & 42.85  & 44.36  \\
\bottomrule
\end{tabular}%
}
\caption{Performance of DTRNet (360M) with token-choice and expert-choice routing strategies.}
\label{tab:expert_vs_token}
\end{table}

Expert Choice routing assigns tokens to different experts based on a global decision computed over the entire input sequence. This approach works well during training, where all tokens are available at once. However, during autoregressive inference, tokens arrive sequentially and the model does not have access to the full sequence at once. This mismatch between training and inference introduces inconsistency in the routing behavior, leading to performance degradation.

Token Choice routing, on the other hand, makes routing decisions independently for each token. This matches the inference-time scenario, ensuring consistent behavior across training and inference.
To empirically compare the two strategies, we implemented an Expert Choice variant of DTRNet where the top 25\% of tokens are routed through attention, while the rest are skipped. The model architecture remained identical to the Token Choice setup, differing only in the routing mechanism. During training, both variants exhibited similar training losses. However, as shown in Table~\ref{tab:expert_vs_token}, the Token Choice model consistently outperforms its Expert Choice counterpart across several benchmarks and achieves a higher overall average accuracy. 

Given its superior alignment with autoregressive inference and stronger empirical results, we adopt Token Choice routing as the routing mechanism in all final DTRNet models.

\subsection{A2. DTRNet Architecture Ablations}

\begin{table}[]
\centering
\small
\resizebox{\linewidth}{!}{%
\begin{tabular}{lcccc}
\toprule
\textbf{Dataset / Metric} & \makecell{\textbf{DTRNet} \\ \textbf{Trilayer}} & \makecell{\textbf{DTRNet} \\ \textbf{Laterhalf}} & \makecell{\textbf{DTRNet} \\ \textbf{6 Transformer}} & \makecell{\textbf{DTRNet} \\ \textbf{Bilayer}} \\
\midrule
WIKI ppl (↓)           & 31.75 & 31.25 & 30.78  & 30.68 \\
LMBD ppl (↓)            & 64.00 & 63.96 & 57.77  & 56.52 \\
LMBD Acc (↑)            & 29.19 & 29.23 & 29.81  & 31.22 \\
ARC-C Acc (↑)           & 23.04 & 22.87 & 23.29  & 23.46 \\
ARC-E Acc (↑)           & 55.56 & 54.71 & 57.11  & 56.06 \\
BQ Acc (↑)           & 61.62 & 57.46 & 58.87  & 59.39 \\
HLSWG Acc (↑)       & 31.21 & 31.36 & 31.75  & 31.73 \\
PIQA Acc (↑)            & 66.05 & 65.45 & 65.07  & 66.38 \\
T-MMLU Acc (↑)          & 31.79 & 30.34 & 28.13  & 32.51 \\
WG Acc (↑)      & 51.14 & 51.93 & 51.38  & 54.14 \\
\midrule
AVG Acc (↑)             & 43.70 & 42.92 & 43.18  & 44.36 \\
\bottomrule
\end{tabular}%
}
\caption{Comparison of four DTRNet architectural variants (360M)}
\label{tab:dtrnet_360m_variants}
\end{table}

We explore multiple architectural configurations for integrating DTRNet blocks into Transformer models. In all variants, the first and last layers are standard Transformer layers. The main configurations are as follows:
\begin{itemize}
    \item \textbf{DTRNet-BiLayer}: Inserts one DTRNet block after each Transformer block. (T-D-T-D-...-D-T)
    \item \textbf{DTRNet-TriLayer}: Inserts two DTRNet blocks after each Transformer block. (T-D-D-T-D-D-...-T)
    \item \textbf{DTRNet-LaterHalf}: The first half of the model consists of full Transformer layers, while the second half consists of DTRNet blocks. (T-T-T-...-D-D-...-D)
    \item \textbf{DTRNet-6Transformer}: A hybrid design where the first two, middle two, and final two layers are full Transformers (6 layers in total), and the remaining layers are DTRNet.
\end{itemize}

Table~\ref{tab:dtrnet_360m_variants} compares four DTRNet architectural configurations on standard language understanding tasks. Among these, DTRNet-BiLayer achieves the highest average accuracy (44.36), followed closely by DTRNet-TriLayer (43.70), outperforming both the LaterHalf (42.92) and 6-Transformer (43.18) variants. This trend is consistent across both perplexity and accuracy metrics, where BiLayer achieves the best performance on most benchmarks, including WIKI, LMBD, ARC-C, BoolQ, PIQA, and Winogrande.

These results support our analysis: token representations in adjacent Transformer layers exhibit high cosine similarity. As a result, selectively skipping attention computation for similar tokens across neighboring layers introduces minimal degradation in contextual understanding. Architectures like BiLayer and TriLayer, which interleave DTRNet blocks with dense layers, exploit this redundancy effectively, preserving performance while reducing computational overhead. In contrast, LaterHalf and 6-Transformer configurations separate DTRNet blocks more coarsely or cluster dense layers, making it harder to leverage local token redundancy, thereby leading to lower performance. This confirms that fine-grained sparsity across adjacent layers offers better performance.

\subsection{A3. Effect of Skipping All Attention}

\begin{table}[h]
\centering
\small
\resizebox{\linewidth}{!}{%
\begin{tabular}{lccc}
\toprule
\textbf{Dataset / Metric} & \makecell{\textbf{SmolLM} \\ \textbf{360M}} & \makecell{\textbf{DTRNet} \\ \textbf{Bilayer}} & \makecell{\textbf{DTRNet} \\ \textbf{Skip}} \\
\midrule
WIKI ppl (↓)              & 30.18  & 30.68  & 30.95 \\
LMBD ppl (↓)              & 52.59  & 56.52  & 61.16 \\
LMBD Acc (↑)              & 30.93  & 31.22  & 28.97 \\
ARC-C Acc (↑)             & 24.32  & 23.46  & 23.04 \\
ARC-E Acc (↑)             & 55.72  & 56.06  & 56.27 \\
BQ Acc (↑)             & 59.88  & 59.39  & 54.10 \\
HLSWG Acc (↑)         & 31.80  & 31.73  & 31.28 \\
PIQA Acc (↑)              & 65.72  & 66.38  & 64.80 \\
T-MMLU Acc (↑)            & 32.62  & 32.51  & 31.76 \\
WG Acc (↑)        & 52.88  & 54.14  & 49.80 \\
FLOPs Ratio               & 1.00   & 0.85   & 0.81   \\
\midrule
AVG Acc (↑)      & 44.23  & 44.36  & 42.50 \\
\bottomrule
\end{tabular}%
}
\caption{Comparison of SmolLM, DTRNet Bilayer and DTRNet Skip (360M)}
\label{tab:dtrnet_skip}
\end{table}

To assess the necessity of routing tokens through the quadratic attention path, we introduce a variant named DTRNet-Skip, where no tokens are passed to the standard multi-head attention in any DTRNet layer. Instead, all tokens are directly forwarded through the linear path. Importantly, the architecture remains identical to the DTRNet-BiLayer configuration, alternating Transformer and DTRNet blocks, with the only difference being that the DTRNet routers are configured to skip all tokens from attention.

As shown in Table~\ref{tab:dtrnet_skip}, although DTRNet-Skip reduces FLOPs by bypassing the most expensive quadratic operation entirely, this comes at a clear cost to performance. The average accuracy drops from 44.36 (DTRNet-BiLayer) to 42.50, and the model underperforms across most tasks. Furthermore, perplexity increases on both WIKI and LAMBADA.

These results confirm that completely avoiding attention degrades model quality. While MLP paths can process and transform information, they lack the global context modeling provided by attention. This experiment demonstrates that selectively passing a small fraction of important tokens to the attention mechanism is essential for maintaining competitive performance while still achieving computational efficiency.

\subsection{A4. MoD and D-LLM original variants}

\begin{table}[h]
\centering
\small
\resizebox{\linewidth}{!}{%
\begin{tabular}{lccccc}
\toprule
\textbf{Dataset / Metric} & \makecell{\textbf{D-LLM} \\ \textbf{(0.55)}} & \makecell{\textbf{D-LLM} \\ \textbf{(0.85)}} & \makecell{\textbf{MoD} \\ \textbf{(k=0.125)}} & \makecell{\textbf{MoD} \\ \textbf{(k=0.7)}} & \makecell{\textbf{DTRNet} \\ \textbf{Bilayer}} \\
\midrule
WIKI ppl (↓)              & 35.13  & 32.82  & 33.80  & 32.16  & 30.68 \\
LMBD ppl (↓)           & 83.61  & 63.31  & 63.10  & 50.00  & 56.52 \\
LMBD Acc (↑)           & 27.13  & 30.12  & 28.97  & 31.48  & 31.22 \\
ARC-C Acc (↑)             & 22.70  & 22.18  & 21.50  & 22.27  & 23.46 \\
ARC-E Acc (↑)             & 52.27  & 54.80  & 51.98  & 54.46  & 56.06 \\
BQ Acc (↑)             & 60.12  & 60.55  & 57.77  & 61.74  & 59.39 \\
HLSWG Acc (↑)         & 30.42  & 31.32  & 30.95  & 31.17  & 31.73 \\
PIQA Acc (↑)              & 64.47  & 63.33  & 63.82  & 64.80  & 66.38 \\
T-MMLU Acc (↑)            & 30.79  & 31.97  & 31.76  & 27.72  & 32.51 \\
WG Acc (↑)        & 50.59  & 48.62  & 50.20  & 51.46  & 54.14 \\
FLOPs Ratio               & 0.56   & 0.85   & 0.55   & 0.85   & 0.85 \\
\midrule
AVG Acc (↑)      & 42.31  & 42.86  & 42.12  & 43.14  & 44.36 \\
\bottomrule
\end{tabular}%
}
\caption{Comparison of original MoD, D-LLM variants with our increased FLOPs variants, and DTRNet Token Choice (360M)}
\label{tab:dtrnet_vs_mod_dllm}
\end{table}

To ensure a fair comparison with DTRNet, we also reimplemented and evaluated the best-performing configurations of MoD and D-LLM as reported in their original papers ($k=0.125$ for MoD, and accelaration rate $0.55$ for D-LLM), using identical model size (360M), training data, and evaluation protocol. These original variants operate at lower FLOPs ratios because they route fewer tokens through the Transformer layer.

As shown in Table~\ref{tab:dtrnet_vs_mod_dllm}, while these variants are efficient, they suffer from reduced performance across most benchmarks compared to their higher FLOPs counterparts. This degradation highlights a fundamental trade-off in sparse routing: reducing computation by skipping too many tokens can hurt model quality. To allow for a more fair comparison, in our main results (Table~\ref{tab:dtrnet_main_table}), we increased the FLOPs budget for both MoD and D-LLM by routing more tokens ($k=0.7$ for MoD, and accelaration rate $0.85$ for D-LLM). These higher FLOPs versions show improved performance, confirming that performance improves with increased token routing. However, even under matched compute, DTRNet consistently outperforms both MoD and D-LLM, demonstrating the effectiveness of our learned token selection strategy.

\subsection{A5. Value and Output Projections Are Needed for Skipped Tokens}

\begin{table}[h]
\centering
\small
\resizebox{\linewidth}{!}{%
\begin{tabular}{lcc}
\toprule
\textbf{Dataset / Metric} & \makecell{\textbf{DTRNet} \\ \textbf{(Bilayer)}} & \makecell{\textbf{DTRNet} \\ \textbf{(Bilayer w/o vo)}} \\
\midrule
WIKI ppl (↓)              & 30.68  & 32.28 \\
LMBD ppl (↓)           & 56.52  & 49.97 \\
LMBD Acc (↑)           & 31.22  & 30.97 \\
ARC-C Acc (↑)             & 23.46  & 22.70 \\
ARC-E Acc (↑)             & 56.06  & 55.35 \\
BQ Acc (↑)             & 59.39  & 50.55 \\
HLSWG Acc (↑)         & 31.73  & 31.50 \\
PIQA Acc (↑)              & 66.38  & 64.47 \\
T-MMLU Acc (↑)            & 32.51  & 28.67 \\
WG Acc (↑)        & 54.14  & 51.78 \\
\midrule
AVG Acc (↑)      & 44.36  & 41.99 \\
\bottomrule
\end{tabular}%
}
\caption{Comparison of DTRNet Bilayer with and without value and output projections under matched hyperparameters (360M)}
\label{tab:dtrnet_vo_ablation}
\end{table}

In DTRNet, tokens routed to the bypass path skip the quadratic attention computation but are still passed through the value ($W^V$) and output ($W^O$) projections before the MLP. This operation is equivalent to a form of \emph{self-attention without interaction}, where a token attends only to itself. By applying $W^V W^O$, the token’s representation is transformed in the same space as tokens from the attention path, maintaining consistency across both routes.

Table~\ref{tab:dtrnet_vo_ablation} shows that removing these projections leads to a clear performance drop across benchmarks (average accuracy falling from 44.36 to 41.99). This confirms that even for skipped tokens, passing them through $W^V$ and $W^O$ enables richer updates and better representational alignment, resulting in improved overall performance compared to simply forwarding them unchanged to the MLP.

\end{document}